\documentclass{article} % For LaTeX2e
\usepackage{iclr2025_conference,times}

\usepackage{amsmath,amsfonts,bm}

\def\eqref#1{equation~\ref{#1}}
\def\1{\bm{1}}

\DeclareMathAlphabet{\mathsfit}{\encodingdefault}{\sfdefault}{m}{sl}
\SetMathAlphabet{\mathsfit}{bold}{\encodingdefault}{\sfdefault}{bx}{n}

\usepackage{hyperref}
\usepackage{url}
\usepackage{graphicx}
\usepackage{booktabs}
\usepackage{multirow}
\usepackage{amsmath}
\usepackage{amssymb}
\usepackage{mathtools}
\usepackage{xspace}
\usepackage{enumitem}
\usepackage{algorithm}
\usepackage{algpseudocode}
\usepackage{xcolor}

\newcommand{\kap}{\textsc{KAP}\xspace}
\newcommand{\graphspec}{\textsc{GraphSpec}\xspace}

\newcommand{\kv}{\textsc{KV}\xspace}

\newcommand{\vllm}{\textsc{vLLM}\xspace}
\newcommand{\topthree}{Top-3\xspace}
\newcommand{\ksrcgap}{knowledge selection--runtime consumption gap\xspace}

\newcommand{\Aplan}{\mathcal{A}}

\newcommand{\Prior}{\pi}

\newcommand{\full}{\mathrm{full}}

\title{\kap: Bridging the Knowledge Selection--Runtime Consumption Gap in LLM~Systems}

\author{Shuo Wang \quad Fang Xi\thanks{Corresponding author.} \quad
Wenyuan Huang \quad Qing Wang \quad Junming Su\\
\multicolumn{1}{c}{QiYuanLab}\\
\multicolumn{1}{c}{Beijing, China}\\
\multicolumn{1}{c}{\texttt{\{wangshuo,xifang,huangwenyuan,sujunming\}@qiyuanlab.com}}\\
\multicolumn{1}{c}{\texttt{d202510468@xs.ustb.edu.cn}}}

\iclrpreprintcopy

\begin{document}

\maketitle

\begin{abstract}
Modern LLM systems increasingly rely on knowledge-selection processes
that produce high-value structured priors, such as ranked evidence, graph
topology, multimodal alignment, and confidence signals.
Yet LLM serving remains fundamentally oblivious to
this rich structure: once such signals are serialized into a prompt, the
backend observes only a flat token sequence, forcing dense and uniform
consumption of the full key-value (\kv) state during decoding. We term this
architectural mismatch the Knowledge Selection--Runtime Consumption
(\textsc{KSRC}) gap: richer contexts enlarge the full-prompt \kv footprint and
decode-time memory traffic, increasing latency and degrading throughput even
when reasoning depends on only a small fraction of the context.
To bridge the gap, we propose \emph{Knowledge Access Planning} (\kap), a
paradigm-shifting execution abstraction that elevates structured knowledge
priors from passive prompt-construction hints into first-class physical
execution artifacts. \kap establishes a universal intermediate representation
(IR)---the runtime access plan---which compiles structured knowledge signals to
govern physical \kv access without altering logical prompt
semantics, model weights, or training procedures. Through this IR, \kap shifts
LLM serving from token-aware context consumption to plan-driven,
knowledge-aware runtime consumption.
We instantiate \kap with \graphspec, a compiler--executor realization
connecting structured knowledge selection to an LLM serving backend.
We derive a phase-boundary model for the positive-speedup regime of
plan-guided execution. Across 4K--128K
long-context QA workloads, \graphspec maintains answer quality comparable to
full-context decoding while decoupling physical \kv consumption from prompt
length, reducing proposal-time \kv access to 5.5\% of source \kv state at
128K, and fundamentally shifting the scaling trajectory of long-context
generation.
\end{abstract}

% =============================================================================
\section{Introduction}
\label{sec:intro}

\begin{figure}[t]
    \centering
    \includegraphics[width=\linewidth]{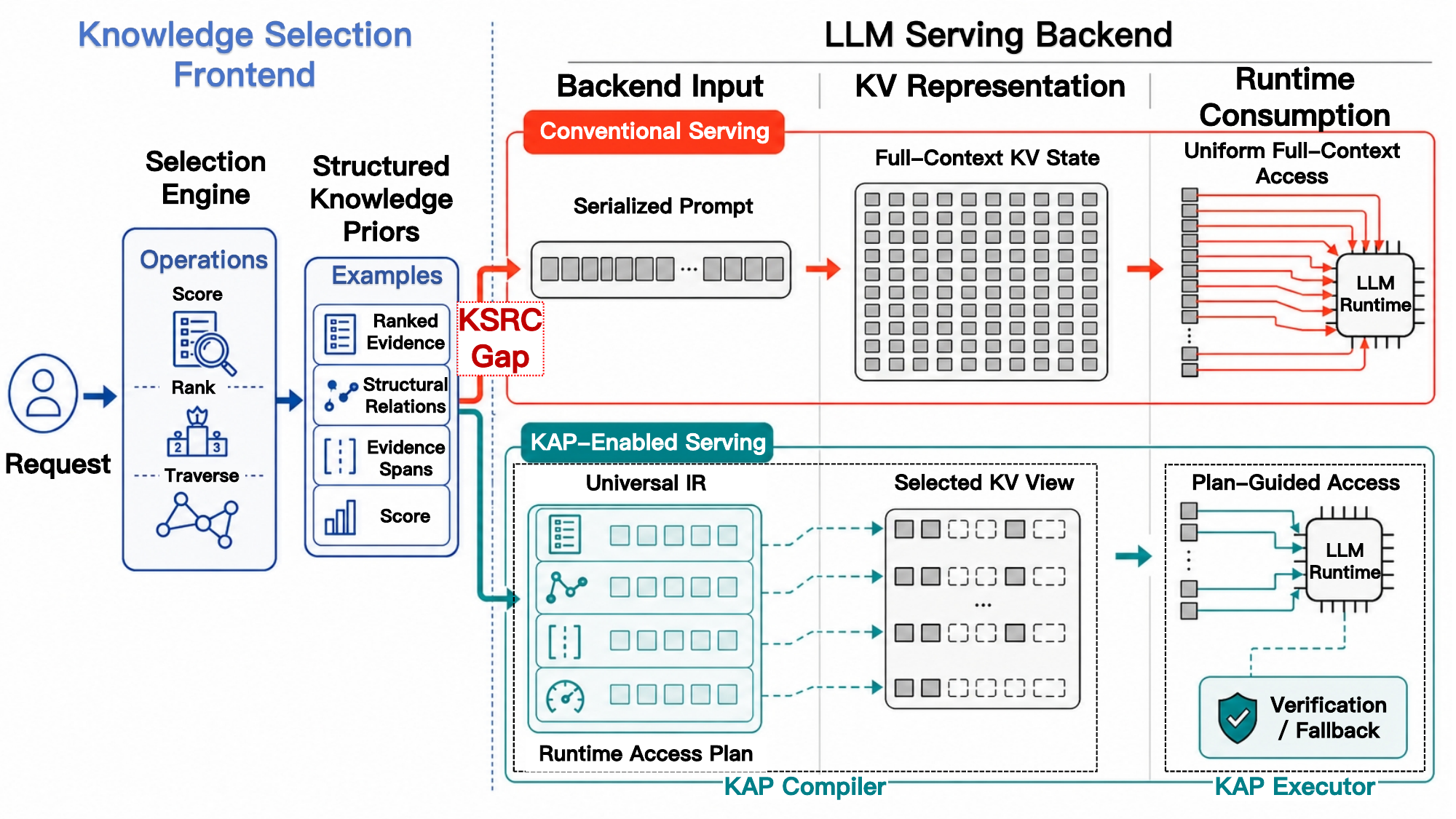}
    \caption{
    \textbf{The \textsc{KSRC} gap and \kap-enabled LLM serving.}
    A knowledge-selection frontend produces structured knowledge priors.
    Conventional serving exposes only a serialized prompt to the LLM serving
    backend and uniformly consumes full-context \kv state, creating the
    \textsc{KSRC} gap. \kap instead carries these priors across the boundary
    through a runtime access plan---a universal IR that determines a selected-\kv
    view and enables plan-guided access with verification or fallback.
    }
    \label{fig:kap_overview}
\end{figure}

\textbf{The Knowledge Selection--Runtime Consumption Gap.}
Modern LLM systems increasingly depend on upstream processes that select,
structure, and prioritize external knowledge. Retrieval-augmented generation
is a prominent instance: modern pipelines combine dense retrieval, reranking,
and structured reasoning frameworks such as GraphRAG
\citep{edge2024from} and HippoRAG \citep{gutierrez2024hipporag} to rank
evidence, traverse graph structure, localize supporting spans, and estimate
confidence \citep{lewis2020retrieval,karpukhin2020dense}. Such
knowledge-selection frontends increasingly operate as semantic planning
systems, yet the LLM serving backend receives only their serialized residue.
Prompt serialization collapses these high-dimensional
knowledge-selection decisions into a one-dimensional token stream, leaving the
backend oblivious to the semantic structure already resolved upstream.
Physical execution is consequently governed by token layout rather than
knowledge importance, wasting frontend semantic computation and backend memory
bandwidth on undifferentiated context. We term the resulting architectural
mismatch the Knowledge Selection--Runtime Consumption Gap: knowledge selection
determines what matters, while the runtime executes as if the full serialized
context must be consumed uniformly. As knowledge-selection processes become
more capable, the gap becomes more pronounced. Richer knowledge inputs,
including graph expansions, larger evidence sets, multimodal context, longer
supporting histories, and other structured signals, may improve reasoning
quality, but they also increase serialized context length, \kv-cache traffic,
decoding latency, and serving cost \citep{pope2022efficient,kwon2023efficient},
often forcing practitioners to discard useful information simply to satisfy
runtime constraints. Knowledge selection has become structure-aware, while
LLM serving remains token-aware.

\textbf{Knowledge Access Planning.}
To bridge the \ksrcgap, we introduce \kap, a general paradigm for plan-driven
knowledge execution. Rather than allowing frontend decisions to terminate at
prompt serialization, \kap organizes runtime knowledge consumption around a
universal intermediate representation (IR)---the runtime access plan---that
decouples logical prompt semantics from physical \kv-cache access. As illustrated in Figure~\ref{fig:kap_overview}, \kap bridges the
\textsc{KSRC} gap through a compiler--executor division of responsibility:
the \kap compiler translates high-dimensional knowledge priors into this IR,
and the \kap executor interprets it to govern physical access within the LLM
serving backend. Together, the compiler and executor shift LLM serving from token-aware context consumption to plan-driven runtime knowledge consumption without modifying the knowledge-selection frontend, language model, or logical prompt.

The architectural scope of the IR extends beyond any single knowledge source.
Because runtime access plans separate frontend semantic units from backend
runtime objects, \kap is independent of any particular modality, model
architecture, or LLM serving substrate. Structured signals such as textual
relevance, graph centrality, multimodal alignment, agent-memory salience, and
source confidence can be compiled into a common execution representation,
while backend-specific executors translate that representation into concrete
runtime actions. Rather than treating larger context windows as a mandate for
proportionally larger physical consumption, \kap makes knowledge consumption
an explicit execution-planning problem.

We argue that bridging the \ksrcgap initiates a fundamental shift in how
long-context LLM systems should consume knowledge. Expanding context windows
toward million-token scales cannot by itself resolve the LLM serving bottleneck
if physical consumption continues to grow with serialization. The future of
scalable reasoning therefore lies in knowledge-selection--execution co-design,
where frontend selectivity and backend consumption are optimized against a
shared systems objective. Section~\ref{sec:cost_analysis} formalizes the coupling
through a phase boundary, providing analytical and empirical grounding for a
new scaling principle: as knowledge selection becomes more selective while
preserving execution fidelity, the physical cost of LLM serving can, and should,
scale sublinearly with context length.

\textbf{GraphSpec as a concrete instantiation.}
To demonstrate the practicality of the \kap paradigm, we instantiate it as
\graphspec, with Graph-RAG serving as the knowledge-selection frontend and
\vllm as the LLM serving backend. \graphspec treats graph-derived frontend
outputs as structured knowledge priors, compiles them into runtime access
plans, and lets the backend consume knowledge according to these plans while
retaining full-context verification.

The instantiation serves as a reference implementation of \kap: graph-derived
priors are no longer used only to construct prompts, but are preserved as
executable guidance for runtime knowledge consumption. \graphspec represents
one concrete realization of the \kap executor; \kap itself does not prescribe
a particular cache organization, access mechanism, or verification strategy.
Section~\ref{sec:graphspec} describes how \graphspec materializes the
abstraction using \vllm runtime objects, and Section~\ref{sec:experiments}
evaluates its serving efficiency and answer quality.

\textbf{Contributions.} Our main contributions are:
\begin{itemize}[leftmargin=*]
    \item \textbf{Problem formulation:} We identify and formalize the
    \ksrcgap as a previously underexplored architectural gap in LLM systems
    that consume externally supplied knowledge. Modern knowledge-selection
    frontends expend substantial computation to estimate, structure, and
    prioritize task-relevant knowledge,
    but prompt serialization prevents these decisions from remaining actionable
    inside the LLM serving backend. The result is a fundamental scaling misalignment: physical serving cost grows with serialized context length rather than with the knowledge that actually drives reasoning.

    \item \textbf{Knowledge Access Planning paradigm:} We introduce \kap, a
    paradigm that shifts LLM serving from passive prompt serialization to
    plan-driven runtime knowledge consumption. At its
    core, \kap establishes the runtime access plan as a universal intermediate
    representation that translates structured knowledge-selection decisions
    into backend physical-access policies while decoupling logical prompt
    semantics from physical \kv consumption.

    \item \textbf{GraphSpec instantiation:} We build \graphspec as a concrete
    reference implementation of \kap, with Graph-RAG serving as the
    knowledge-selection frontend and \vllm as the LLM serving backend. By
    connecting frontend knowledge priors to the LLM serving runtime through
    runtime access plans, \graphspec demonstrates that \kap is not only a
    conceptual paradigm but a deployable serving abstraction that can be
    realized within an existing LLM serving stack.

    \item \textbf{Phase-boundary analysis:} We derive a phase-boundary model
    that quantifies how knowledge-guided selectivity and runtime acceptance jointly
    determine the positive-speedup region. The resulting conditional boundary
    provides an analytical criterion for knowledge-selection--execution
    co-design across the frontend, the \kap compiler, and the LLM serving backend.

    \item \textbf{Empirical evaluation:} We conduct a controlled empirical
    evaluation on long-context QA workloads with up to 128K-token contexts. At
    128K, \graphspec exposes only 5.5\% of the source \kv state to the proposal
    path while maintaining answer quality comparable to full-context decoding,
    with a corresponding 1.19$\times$ decode-throughput speedup. The observed
    behavior is consistent with the predicted phase-boundary trend, supporting both
    the \kap paradigm and its \graphspec realization.
\end{itemize}

\section{Knowledge Access Planning}
\label{sec:kap}

\kap is organized around a universal intermediate representation: the
\emph{runtime access plan}. Rather than defining \kap by a particular
sparse-access mechanism, cache layout, or decoding procedure, this runtime IR
carries structured knowledge selection across the prompt boundary and into LLM
serving. It provides a common representation through which frontend semantic
decisions can be compiled into physical-access policies interpreted by the LLM
serving backend. We first define the IR schema and its invariants, and then
present the compiler--executor architecture that produces and consumes it.

\subsection{Runtime Access Plan as a Universal IR for LLM Serving}
\label{subsec:runtime_access_plan}

A runtime access plan acts as the execution contract between knowledge selection
and LLM serving. Knowledge-selection frontends reason over semantic objects
such as passages, evidence spans, graph nodes, and relevance signals, whereas
LLM serving runtimes execute over physical objects such as token spans, \kv
objects, cache blocks, attention regions, and scheduling metadata. The plan bridges these
abstraction levels by specifying how frontend-selected knowledge is
represented, grounded in the logical prompt, mapped to backend runtime objects,
accessed during decoding, and validated before outputs are committed. The
contract separates the logical context that defines model conditioning from the
physical access pattern used for efficient execution.

We represent the universal IR as:
\begin{equation}
    \Aplan =
    (U, M, pos, \rho_{\mathrm{access}}, \rho_{\mathrm{verify}}),
    \label{eq:access_plan}
\end{equation}
where $U$ contains selected knowledge units, $M$ binds those units to runtime
objects and their physical organization, $pos$ preserves their logical
positions, and $\rho_{\mathrm{access}}$ and $\rho_{\mathrm{verify}}$ encode
access and verification behavior. These fields form the common semantic
contract between compiler-specific priors and executor-specific runtime
objects.

To fulfill its role, a runtime access plan answers five complementary design
questions. \textbf{What knowledge should participate in execution?} Selected
knowledge units identify the prioritized evidence retained under a
runtime budget. \textbf{Where is the selected knowledge located at runtime?}
Runtime object mappings bind logical evidence to objects exposed by the LLM
serving runtime, including token spans, \kv objects, cache blocks, attention
regions, and other runtime metadata. \textbf{How can runtime optimization preserve model
semantics?} Logical positions ensure that selective or reorganized \kv access
does not alter the positional semantics perceived by the model
\citep{su2024roformer,liu2024lost}. \textbf{How should selected knowledge be
consumed during decoding?} The runtime access policy specifies how the executor
exposes selected objects during execution. \textbf{How is execution correctness
guarded?} The verification policy defines how selective execution is verified
or reverted before outputs are committed.

Collectively, these five dimensions define the plan schema exposed by the \kap
compiler to the \kap executor. The schema makes selected knowledge actionable
at runtime, but a valid execution abstraction must also preserve the invariants
that keep selective access faithful to selection intent, model semantics, and
output correctness. We therefore require runtime access plans to satisfy three
cross-field requirements. \textbf{Prior preservation} ensures that frontend
knowledge signals are not lost at the prompt boundary, but remain available
as runtime guidance throughout execution. \textbf{Logical-position
preservation} separates physical \kv organization from the logical ordering and
positional semantics perceived by the model, so runtime optimization does not
rewrite the conditioning context. \textbf{Verification-aware execution} makes
selective access a guarded execution path: outputs produced under plan-guided
access are committed only after reconciliation with full-context verification
or fallback. Together, these requirements distinguish runtime access plans from
ordinary frontend metadata: they preserve the semantic structure produced by
knowledge selection while exposing the execution structure needed for runtime
optimization.

The resulting plan abstraction connects knowledge selection to runtime
execution without prescribing a particular cache organization, access
mechanism, verification strategy, or LLM serving backend.
These requirements also clarify that an effective plan is not simply the one
that exposes the least runtime state, but the one that balances selective
execution with the fidelity needed for verified generation. Section~\ref{sec:cost_analysis}
formalizes the quality-efficiency trade-off through a phase-boundary analysis,
while the next subsection maps the plan abstraction onto the \kap
compiler--executor architecture.

\subsection{\kap compiler--executor architecture}
\label{subsec:kap_paradigm}

\begin{figure}[t]
    \centering
    \includegraphics[width=\linewidth,trim={0pt 34pt 0pt 22pt},clip]{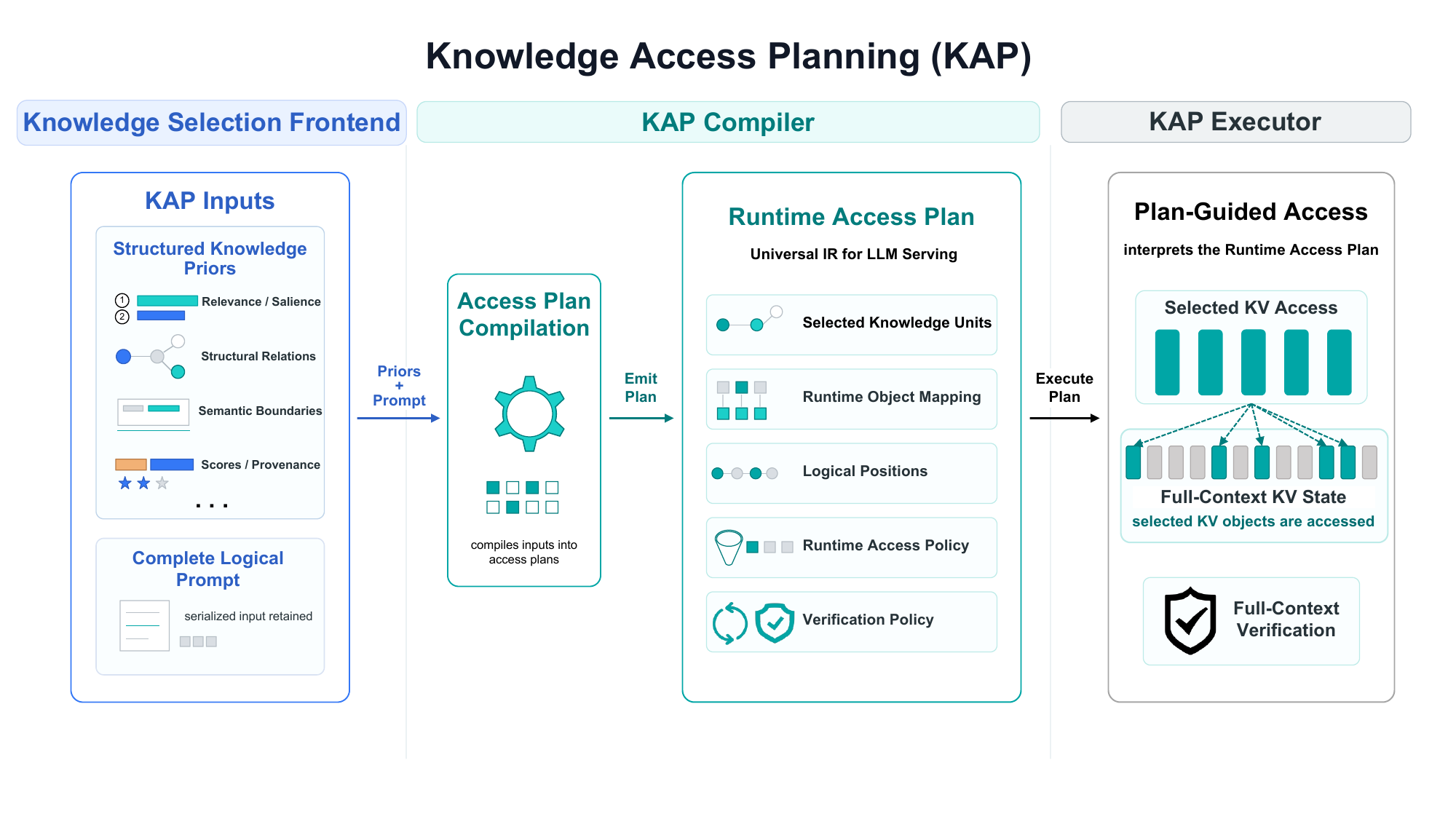}
    \caption{
    \textbf{\kap compiler--executor architecture.}
    The \kap compiler takes structured knowledge priors together with the complete
    logical prompt and emits a runtime access plan, the universal IR for LLM
    serving. The plan encodes selected knowledge units, runtime object mappings,
    logical positions, runtime access policy, and verification policy. The \kap
    executor interprets the plan to perform plan-guided selected-\kv access over
    full-context \kv state and applies the specified verification policy,
    illustrated here through full-context verification.
    }
    \label{fig:kap_architecture}
\end{figure}

The \kap architecture operationalizes the runtime access plan through an
explicit compiler--executor separation. As illustrated in
Figure~\ref{fig:kap_architecture}, the knowledge-selection frontend retains its
existing role: it selects and prioritizes task-relevant knowledge, producing
structured knowledge priors together with the complete logical prompt. The \kap compiler
interprets these artifacts under a runtime budget and emits a runtime access
plan that identifies selected knowledge, binds it to runtime objects, records
logical positions, and specifies access and verification policies. The \kap
executor interprets the emitted plan as runtime actions, exposing selected \kv objects
during decoding and invoking full-context verification or fallback before
outputs are committed.

The division of responsibility realizes the requirements established in the
preceding subsection. Compilation preserves knowledge priors and carries
logical positions into backend-visible mappings; execution applies the access
policy while enforcing the verification policy. The complete logical prompt
therefore remains the model's conditioning context even when physical \kv
access is selective. By separating knowledge selection, plan compilation, and
plan-guided execution, \kap makes selected knowledge actionable within LLM
serving without requiring changes to the knowledge-selection frontend or the
base language model. More fundamentally, the compiler--executor separation
defines \kap as a model-, modality-, and framework-independent systems
abstraction: heterogeneous priors from multimodal contexts, agent histories,
graph-structured knowledge, and other sources can be compiled into a common
runtime access-plan IR, while backend-specific executors realize the same IR
across diverse cache organizations, attention mechanisms, and inference
substrates. \kap thereby provides an extensible foundation for
knowledge-selection--execution co-design, allowing application-specific
frontends and backend execution mechanisms to evolve independently through a
common executable representation. Section~\ref{sec:graphspec} shows how
\graphspec instantiates the architecture for a Graph-RAG frontend and the
\vllm serving backend.

% =============================================================================

\section{GraphSpec: Instantiating \kap with a Graph-RAG Frontend and \vllm Backend}
\label{sec:graphspec}

\begin{figure}[t]
    \centering
    \includegraphics[width=\linewidth,trim={0pt 50pt 0pt 34pt},clip]{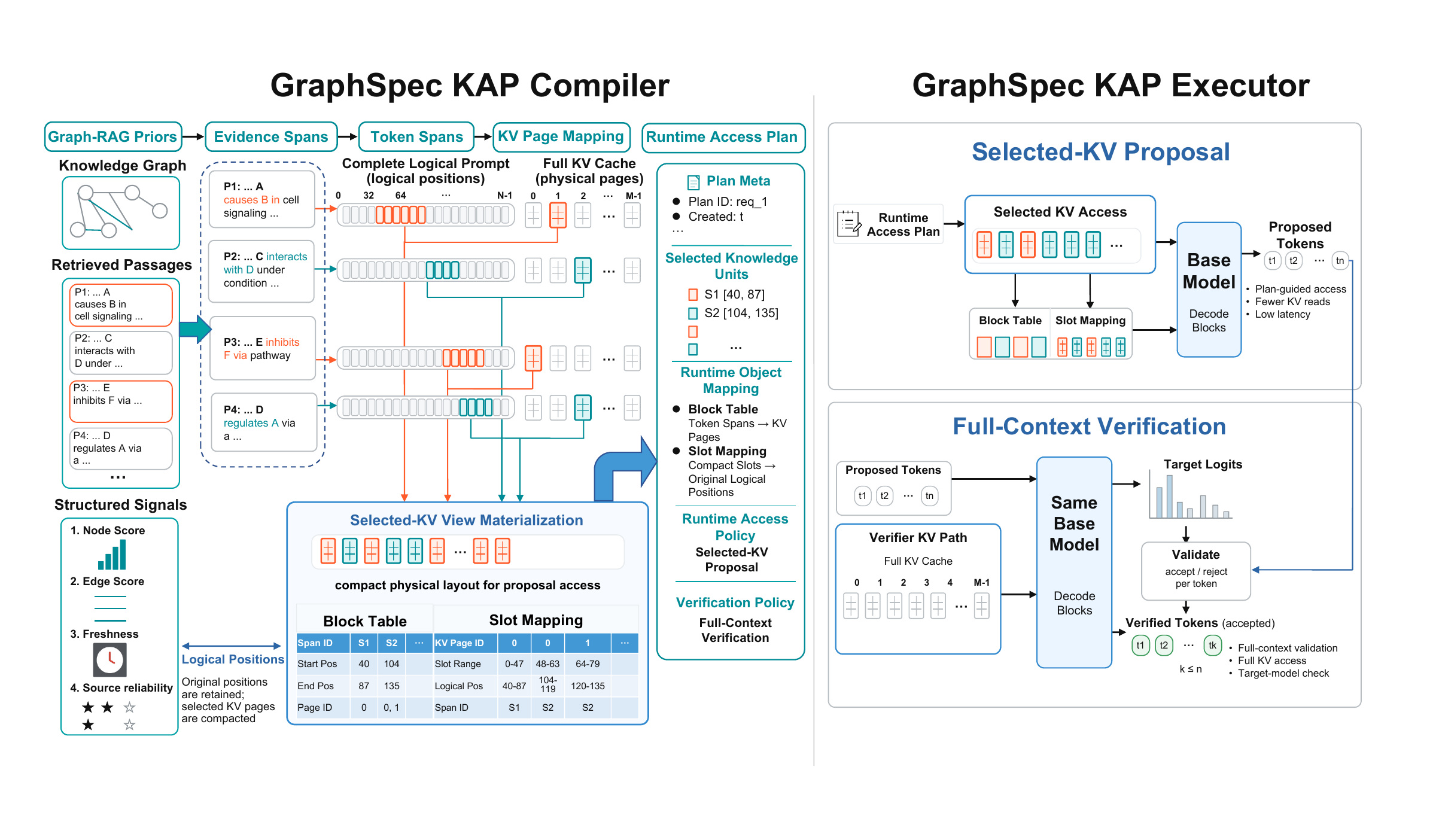}
    \caption{
    \textbf{\graphspec realization of the \kap compiler--executor architecture.}
    Compiler: the GraphSpec \kap compiler grounds Graph-RAG priors in evidence and
    token spans, maps selected spans to paged-\kv runtime objects, materializes a
    logical-position-preserving selected-\kv view, and encodes the corresponding
    mappings and policies in a runtime access plan.
    Executor: the GraphSpec \kap executor interprets the plan through selected-\kv
    proposal, reconciles proposed tokens through full-context verification with
    the same base model, and commits the accepted prefix.
    }
    \label{fig:graphspec_design}
\end{figure}

We present \graphspec as a reference implementation of the universal runtime
access plan IR and compiler--executor architecture introduced by \kap in
Section~\ref{sec:kap}. Rather than defining a new knowledge-selection frontend
or serving engine, \graphspec specializes \kap to an integrated Graph-RAG
frontend and \vllm serving backend. The Graph-RAG frontend supplies
graph-derived knowledge priors and the complete logical prompt, while \vllm
provides the execution substrate for
plan-guided \kv access. The runtime access plan carries graph-level selection
decisions into backend runtime behavior.
\graphspec selects one point in the open executor design space: selected-\kv
self-speculation with full-context verification. This execution policy provides
a concrete realization of \kap rather than defining the \kap abstraction
itself.
Figure~\ref{fig:graphspec_design} concretizes the universal IR by mapping each
plan dimension in Section~\ref{subsec:runtime_access_plan} to \vllm runtime
state. \graphspec represents selected knowledge units as budgeted evidence
spans; realizes runtime object mappings and logical positions through
token-span-to-page mappings, block tables, slot mappings, and preserved
position indices; and instantiates the access and verification policies as
selected-\kv proposal and full-context verification. The runtime access plan
therefore remains the knowledge-level IR, whereas the selected-\kv view is its
\vllm-specific proposal-side materialization; the executor coordinates this
physical view with full-context verification before committing tokens.

\subsection{GraphSpec \kap compiler}
\label{subsec:graphspec_compiler}
\label{subsec:graph_prior_construction}

\paragraph{Graph-RAG priors.}
GraphSpec treats the Graph-RAG frontend as a source of graph-derived knowledge
priors rather than as part of the serving runtime. In our implementation, the
priors are produced by a hybrid Graph-RAG retrieval procedure that combines vector
retrieval, reranking, graph node scoring, and PPR-based graph expansion to
produce candidate evidence. Each retrieved graph unit may carry frontend
metadata such as node type, passage provenance, graph neighborhood, anchor
location, or other annotations useful for runtime access.
Graph-RAG thus realizes one point in a broader \kap frontend space:
application-specific knowledge structures enter LLM serving as compilable
priors rather than backend-specific execution logic.

\paragraph{Evidence-to-runtime compilation.}
The compiler converts these priors into a backend-executable plan in four
stages. It first grounds graph-derived evidence in source passages and aligns
the evidence to token spans in the complete logical prompt. It then performs
budget-aware semantic selection over evidence objects rather than \kv pages.
Next, it resolves the selected spans to \kv pages and materializes a compact
physical view while preserving logical positions and constructing the block
tables and slot mappings required by the executor. Finally, it emits a runtime
access plan that references these runtime objects and specifies selected-\kv
proposal and full-context verification policies. Algorithm~\ref{alg:graphspec_compiler}
formalizes the GraphSpec-specific compilation path.
The staged translation exposes the portability of the universal IR: semantic
selection remains application-defined, while physical realization is
specialized by the backend compiler.

\begin{algorithm}[t]
    \caption{\textbf{GraphSpec \kap compiler.} Graph-RAG-specific access-plan compilation.}
    \label{alg:graphspec_compiler}
    \begin{algorithmic}[1]
        \Procedure{GraphSpecCompile}{$\Prior_{\mathrm{graph}}, X, \mathcal{P}_{\mathrm{KV}}, B, \rho_{\mathrm{verify}}$}
        \Statex \textbf{Input:} Graph-RAG priors $\Prior_{\mathrm{graph}}$, logical prompt $X$, paged-\kv metadata $\mathcal{P}_{\mathrm{KV}}$, budget $B$, verification policy $\rho_{\mathrm{verify}}$
        \Statex \textbf{Output:} runtime access plan $\mathsf{plan}$
        \State $\mathsf{spans} \gets \Call{GroundEvidence}{\Prior_{\mathrm{graph}}, X}$
        \State $\mathsf{tokenSpans} \gets \Call{AlignTokenSpans}{\mathsf{spans}, X}$
        \State $U \gets \Call{SelectEvidenceUnderBudget}{\mathsf{tokenSpans}, \Prior_{\mathrm{graph}}, B}$
        \State $V \gets \Call{ResolveSelectedKVObjects}{U, \mathsf{tokenSpans}, \mathcal{P}_{\mathrm{KV}}}$
        \State $pos \gets \Call{PreserveLogicalPositions}{U, \mathsf{tokenSpans}, X}$
        \State $\mathsf{slots} \gets \Call{AssignCompactSlots}{V}$
        \State $\mathsf{blockTable} \gets \Call{BuildBlockTable}{V, \mathsf{slots}, \mathcal{P}_{\mathrm{KV}}}$
        \State $\mathsf{slotMap} \gets \Call{BuildSlotMapping}{U, \mathsf{tokenSpans}, \mathsf{slots}}$
        \State $\mathsf{view} \gets \Call{BuildSelectedKVView}{V, \mathsf{blockTable}, \mathsf{slotMap}, pos}$
        \State $M \gets \{\mathsf{view}, \mathsf{blockTable}, \mathsf{slotMap}\}$
        \State $\rho_{\mathrm{access}} \gets \Call{BuildSelectedKVProposalPolicy}{\mathsf{view}, B}$
        \State $\mathsf{plan} \gets \Call{EmitRuntimeAccessPlan}{U, M, pos, \rho_{\mathrm{access}}, \rho_{\mathrm{verify}}}$
        \State \Return $\mathsf{plan}$
        \EndProcedure
    \end{algorithmic}
\end{algorithm}

\paragraph{Selected-\kv view materialization.}
The compiler materializes the proposal-side access specification carried by
the runtime access plan as a selected-\kv view over \vllm's paged-\kv cache. As
shown by the selected-\kv-view and
runtime-object-mapping components in Figure~\ref{fig:graphspec_design},
semantic evidence selections are resolved into token-span and \kv-page
mappings, and the selected pages are compacted in access order. Crucially, this
materialization reorganizes only physical \kv placement: the original logical
positions are preserved through the visibility mask, prompt-length metadata,
block tables, and slot mappings consumed by the attention backend. Block tables
resolve plan-visible logical blocks to compact physical pages, while slot
mappings place newly scheduled tokens into their corresponding physical
\kv slots.

During proposal, these metadata expose only the \kv objects selected by the
runtime access plan; during verification, full-context metadata exposes the
complete \kv state. Both paths retain the same logical positions, keeping
proposal and verifier logits aligned under consistent positional semantics.
The verification policy remains associated with the request and invokes
full-context fallback when selected-\kv execution is invalid. The resulting
selected-\kv view therefore exemplifies how the universal plan IR can be
specialized into engine-specific physical state while preserving model
conditioning, a realization pattern that extends beyond \vllm's paged-\kv
organization.

\subsection{GraphSpec \kap executor}
\label{subsec:graphspec_executor}
\label{subsec:sparse_draft_path}
\label{subsec:full_context_verify_resync}
\label{subsec:metadata_cuda_graph}

\paragraph{Selected-\kv proposal.}
The GraphSpec \kap executor interprets the runtime access plan through two
coordinated execution paths. The plan references the materialized selected-\kv
view and its block-table, slot-mapping, positional, and verification metadata;
GraphSpec uses the same base model for proposal generation and verification.
Recent long-context speculative systems reduce proposal cost through shortened
retrieval contexts, quantized draft states, or learned memory-efficient
drafters \citep{chen2025rapid,tiwari2025quantspec,yang2026longspec}.
GraphSpec adopts the same general draft-and-verify structure but derives its
proposal path differently: the base model executes over a selected-\kv view
compiled from frontend knowledge priors through the runtime access plan.
Each proposal step is implemented as a decode-like forward pass with
$q_{\mathrm{prop}}=1$. The executor provides the current proposed token, its
full-sequence logical position, the compact block table, compact slot mapping,
and compact attention metadata constructed by the compiler. The model
parameters are identical to the full-context verifier; only the prompt-side
\kv objects exposed to attention are restricted by the selected \kv view.

As illustrated on the executor side of Figure~\ref{fig:graphspec_design}, the
proposal state is provisional. The access plan determines which prompt-side
\kv pages are visible during candidate generation, but final acceptance and
committed-state advancement remain with the full-context verifier. Proposal
efficiency therefore comes from executing the same model over selected-\kv
access while retaining a full-context verification path.
At the architectural level, this realization shows how a \kap executor can
introduce a specialized knowledge-consumption path alongside an existing
full-context path without changing the underlying model.

\paragraph{Full-context verification.}
After proposing $k$ candidate tokens, GraphSpec verifies them with a
full-context verifier path. The verification input is a short contiguous query
window with $q_{\mathrm{verify}}=k+1$: a verifier-approved carry token that
anchors the window, followed by the $k$ proposed tokens. The verifier uses the
full \kv cache, full-context block table, full-context slot mapping, and
original logical positions.

Let $c$ denote the carry token and let $\hat{\mathbf{x}}_{1:k}$ denote the
proposed token window. The verifier forward is the runtime transition that
both validates proposals and advances the full-context state:
\begin{equation}
    (m, c^{+}, \mathrm{KV}^{\mathrm{full},+})
    =
    \mathcal{V}^{\mathrm{full}}_{\theta}
    \left(
        \mathrm{KV}^{\mathrm{full}},
        c,
        \hat{\mathbf{x}}_{1:k};
        \rho
    \right),
    \label{eq:verify_state_transition}
\end{equation}
where $\rho$ is the verification policy and $m \in \{0,\ldots,k\}$ is the
number of proposed tokens accepted consecutively from the beginning of the
candidate window. For QA-oriented serving, we instantiate $\rho$ as
target-guided tolerant verification: the \topthree rule used in our
implementation accepts a proposed token when it lies within the target model's
top-3 candidates under the full context. The resulting acceptance count $m$ is
the runtime quantity underlying the acceptance term $h$ in
Section~\ref{sec:cost_analysis}.

The verifier transition produces the accepted prefix, the next
verifier-approved carry token $c^{+}$, and the committed full-context state
$\mathrm{KV}^{\mathrm{full},+}$ for the committed sequence. Full-context
verification therefore advances the full-\kv state within the same verifier
transition, eliminating a separate resynchronization pass. As shown by the
verifier \kv path in Figure~\ref{fig:graphspec_design}, selected-\kv proposal
access is reconciled with the full-context state before tokens are committed,
providing the quality guard for selective execution.
Encoding this reconciliation as a plan policy separates physical-access
optimization from output-commitment semantics, allowing executor strategies
to evolve without weakening the correctness path.

\paragraph{Runtime stability and fallback.}
Reducing proposal-time \kv access is necessary but not sufficient for
end-to-end speedup: dynamic metadata, changing tensor shapes, or repeated
\kv allocation can erase the savings from selected-\kv access. At the
implementation level, the selected-\kv proposal path must remain close to
\vllm's optimized uniform decode path: the plan should restrict \kv visibility
without turning decoding into a shape-changing metadata-reconstruction path.
GraphSpec therefore keeps proposal and verification shapes stable whenever
possible \citep{kwon2023efficient}. The proposal path follows a stable
decode-like execution shape in \vllm with compact attention metadata, while the
verifier uses a fixed $q_{\mathrm{verify}}=k+1$ window for a chosen proposal
length $k$.

In practice, GraphSpec stabilizes query lengths, compact \kv capacity, block
structure, block tables, slot mappings, attention metadata shapes, and CUDA
graph dispatch tags. A reusable compact \kv workspace reduces allocation
overhead and makes the proposal path more graph-replay friendly. Consequently,
the block-table and slot-mapping objects in
Figure~\ref{fig:graphspec_design} are performance-critical runtime metadata:
their regularity determines whether selected-\kv access preserves \vllm's
optimized decode execution. These constraints also explain why proposal-side
\kv reduction is not expected to translate linearly into end-to-end
throughput: selected-\kv access reduces one component of the critical path,
while verification and runtime orchestration remain necessary.

Fallback combines runtime feasibility and plan quality. GraphSpec uses
full-context access when the plan exceeds the compact workspace, violates
metadata-stability assumptions, has insufficient graph-prior confidence,
retains too much \kv ($s$ high), or yields low observed acceptance ($h$ low).
These conditions operationalize the quantities underlying the phase boundary
in Section~\ref{sec:cost_analysis}: a plan must satisfy the $h$--$s$ trade-off
while preserving the execution
regularity required by high-performance LLM serving.
Making feasibility and fallback conditions visible to planning turns backend
constraints into compiler inputs, enabling knowledge selection, plan
construction, and physical execution to be optimized as a coordinated system.

% =============================================================================

\section{Cost Model and Phase-Boundary Analysis}
\label{sec:cost_analysis}

A runtime access plan reduces serving cost by exposing only a selected portion
of the available \kv state to a planned execution path. However, selective
access also introduces a quality-efficiency trade-off: removing too much
context may reduce proposal quality, lower verifier acceptance, and increase
the cost paid per committed token. We develop a cost model for plan-guided
execution and derive a phase boundary that separates positive- and
negative-speedup regimes, first for a general proposal--verification
setting, and then specialize the result to \graphspec, where the proposal and
verification paths use the same base model.

\subsection{Cost model and conditional phase boundary}
\label{subsec:cost_model}

Consider a runtime access plan $\Aplan$ produced by the \kap compiler according
to the universal IR in Equation~\ref{eq:access_plan}. Standard full-context
serving corresponds to consuming the full prompt access set
$A_{\mathrm{full}}=\{1,\ldots,L\}$ at each decoding step; runtime
access planning instead exposes a budgeted subset or physical view derived from
$\Aplan$ while preserving full-context verification.

\paragraph{Model assumptions.}
We adopt a memory-bound view of autoregressive decoding
\citep{pope2022efficient,kwon2023efficient}, treat context length as locally
constant within one proposal--verification cycle, and model weight and \kv
reads as effective memory-traffic costs. The analysis abstracts away scheduler,
metadata-construction, synchronization, and kernel-launch overheads; these
factors affect realized speedup but not the idealized break-even condition
derived below. Let $W_{\mathrm{main}}$ denote the effective weight-memory cost
of one full-context main-model decode forward, and let $KV$ denote the cost of
reading the full historical \kv cache. Standard full-context decoding incurs:
\begin{equation}
    \mathrm{Cost}_{\mathrm{full}} \approx W_{\mathrm{main}} + KV .
    \label{eq:full_context_cost}
\end{equation}
A runtime access plan affects the proposal path through two quantities. The
retained-\kv ratio is:
\begin{equation}
    s(\Aplan) =
    \frac{\text{\kv accessed by the proposal path}}
         {\text{\kv accessed by full-context decoding}} ,
    \label{eq:retained_ratio}
\end{equation}
and the proposal acceptance rate is:
\begin{equation}
    h(\Aplan) =
    \frac{\text{accepted proposed tokens}}
         {\text{speculated proposed tokens}} .
    \label{eq:acceptance_rate}
\end{equation}
We write these as $s$ and $h$ when $\Aplan$ is clear.

Consider a planned cycle that proposes $k$ tokens and verifies them with the
full-context main model. Let $W_{\mathrm{prop}}$ denote the per-step
weight-memory cost of the proposal path. The cycle produces $1+kh$ output
tokens in expectation, so standard full-context decoding would spend:
\begin{equation}
    \mathrm{Cost}_{\mathrm{base}}
    =
    (1 + kh)(W_{\mathrm{main}} + KV).
    \label{eq:baseline_equal_output_cost}
\end{equation}
The planned cycle spends:
\begin{equation}
    \mathrm{Cost}_{\mathrm{plan}}
    =
    k(W_{\mathrm{prop}} + sKV)
    +
    (W_{\mathrm{main}} + KV),
    \label{eq:planned_spec_cost}
\end{equation}
where the first term is the proposal cost under plan-guided selected-\kv access
and the second term is full-context verification. A runtime access plan yields
positive speedup when it costs less than standard full-context decoding for the
same expected output:
\begin{equation}
    \mathrm{Cost}_{\mathrm{plan}}
    <
    \mathrm{Cost}_{\mathrm{base}} .
    \label{eq:plan_speedup_condition}
\end{equation}
Substituting Equations~\ref{eq:baseline_equal_output_cost} and~\ref{eq:planned_spec_cost}, we obtain:
\begin{align}
    k(W_{\mathrm{prop}} + sKV) + (W_{\mathrm{main}} + KV)
    &<
    (1 + kh)(W_{\mathrm{main}} + KV) \nonumber \\
    W_{\mathrm{prop}} + sKV
    &<
    h(W_{\mathrm{main}} + KV) \nonumber \\
    (h-s)KV
    &>
    W_{\mathrm{prop}} - hW_{\mathrm{main}} .
    \label{eq:general_phase_boundary}
\end{align}
Equation~\ref{eq:general_phase_boundary} defines the idealized phase boundary
for the proposal--verification class of plan-guided execution. Conditional on
the $h$ and $s$ induced by a plan, it separates the positive- and
negative-speedup regimes under the cost-model assumptions. The left-hand side
is the acceptance-adjusted \kv-access margin, while the right-hand side is the
residual proposal-path cost after accounting for accepted main-model work. A
plan is beneficial only when it reduces retained \kv access enough, and
preserves proposal acceptance enough, to offset proposal-path and full-context
verification costs.

\subsection{GraphSpec specialization and planning implications}
\label{subsec:self_spec_boundary}

GraphSpec is a self-speculative instantiation of \kap: the proposal and
verification paths use the same base model. Therefore,
\begin{equation}
    W_{\mathrm{prop}} = W_{\mathrm{main}} .
\end{equation}
With equal proposal and verification model costs,
Equation~\ref{eq:general_phase_boundary} becomes:
\begin{equation}
    (h-s)KV > (1-h)W_{\mathrm{main}} .
    \label{eq:self_spec_boundary}
\end{equation}
The specialized boundary exposes the basic margin condition behind
GraphSpec-style selected-\kv self-speculation:
\begin{equation}
    h > s .
    \label{eq:h_greater_s}
\end{equation}
The condition is necessary for a positive \kv-access margin, while realized
speedup still depends on the break-even \kv scale below and practical runtime
overheads. If $h \leq s$, the selected-\kv proposal path does not create such a
margin; increasing context length alone cannot move the plan into the
positive-speedup region. When $h>s$, the approximate idealized break-even \kv
scale is:
\begin{equation}
    KV_{\mathrm{crit}}
    \approx
    \frac{(1-h)W_{\mathrm{main}}}{h-s}.
    \label{eq:kv_crit}
\end{equation}
Thus, GraphSpec is naturally long-context oriented: as context length grows,
$KV$ increases, and a plan with high acceptance and low retained-\kv ratio can
cross the phase boundary.

For \kap, the boundary turns access planning into a constrained plan-selection
problem: retain enough critical evidence to keep the proposal path aligned
with the full-context verifier, while removing enough low-value \kv access to
reduce runtime cost. We define the ideal phase slack of a runtime access plan
as:
\begin{equation}
    \Delta_{\mathrm{phase}}(\Aplan)
    =
    h(\Aplan)(W_{\mathrm{main}}+KV)
    -
    \left(W_{\mathrm{prop}}+s(\Aplan)KV\right).
    \label{eq:phase_slack}
\end{equation}
A plan lies in the positive-speedup region when
$\Delta_{\mathrm{phase}}(\Aplan)>0$. The corresponding phase-slack objective is:
\begin{equation}
    \Aplan_{\mathrm{phase}}^{*}
    =
    \arg\max_{\Aplan}
    \Delta_{\mathrm{phase}}(\Aplan),
    \quad
    \text{s.t.}
    \quad
    Q_{\mathrm{task}}(\Aplan) \approx Q_{\mathrm{task}}(\full),
    \label{eq:kap_access_plan_objective}
\end{equation}
where $Q_{\mathrm{task}}$ denotes answer-level task quality. For \graphspec,
$\Delta_{\mathrm{phase}}=(h-s)KV-(1-h)W_{\mathrm{main}}$. Importantly, $h$ and
$s$ are coupled through the quality of the compiled plan. Retaining more
runtime state increases $s$ but may improve $h$ by preserving context that
aligns the proposal path with the full-context verifier; more selective access
lowers $s$ but may reduce $h$. Knowledge-prior quality shapes the trade-off:
informative priors can sustain a given acceptance level at a smaller
retained-\kv ratio, whereas weak or miscalibrated priors require broader
retention or move execution toward the negative-speedup region. The quantity
$h-s$ captures the \kv-side acceptance--retention margin, whereas
$\Delta_{\mathrm{phase}}$ represents the complete idealized break-even slack.
The compiler should therefore optimize phase slack rather than minimize
retained \kv access in isolation. The phase boundary converts
knowledge-selection quality into execution utility: knowledge priors are
valuable not only when they identify semantically relevant evidence, but also
when they induce runtime access plans with favorable break-even behavior. This
establishes knowledge-selection--execution co-design as a quantitative systems
objective rather than a qualitative architectural principle.

The boundary provides a planning criterion rather than a guarantee of proposal
quality. The compiler can vary the access budget and thereby influence $s$,
while $h$ is induced by the resulting plan and verification policy. Observed
acceptance and profiled runtime costs can therefore inform budget adjustment or
fallback when the estimated phase slack becomes unfavorable.

Three practical implications follow. First, the proposal length $k$ cancels out
of the ideal break-even condition, so it does not determine whether a plan can
enter the positive-speedup region; instead, it shapes finite-cycle performance
by amortizing full-context verification and interacting with runtime overheads.
Second, proposal and verification logits must remain comparable, which is why
GraphSpec preserves logical positions even when physical \kv access is
selective or compacted. Third, $h(\Aplan)$ depends on the verification policy.
For the QA-oriented setting evaluated by GraphSpec, we instantiate $h$ as the
\topthree tolerant acceptance rate computed against the full-context verifier:
\begin{equation}
    (h_{\mathrm{top3}} - s)KV
    >
    (1-h_{\mathrm{top3}})W_{\mathrm{main}} .
    \label{eq:top3_boundary}
\end{equation}
Equation~\ref{eq:top3_boundary} instantiates the conditional cost model for
target-guided tolerant verification; it is not an exact
sampling guarantee; answer quality is evaluated empirically against standard
full-context decoding.

% =============================================================================

\section{Experiments}
\label{sec:experiments}

We evaluate \graphspec as an instantiation of \kap for long-context
knowledge-augmented LLM serving through four research questions that organize
the evidence chain from the \ksrcgap diagnosis to \kap's systems behavior,
execution mechanism, and analytical boundary:
\begin{itemize}[leftmargin=*]
    \item \textbf{RQ1:} Does \graphspec decouple physical \kv consumption from
    serialized context growth?
    \item \textbf{RQ2:} Does \graphspec sustain efficiency gains while
    preserving answer quality across context lengths?
    \item \textbf{RQ3:} Does the runtime access plan bridge knowledge selection
    and runtime consumption?
    \item \textbf{RQ4:} Is the observed speedup trend consistent with the
    phase-boundary analysis?
\end{itemize}

\subsection{Setup}
\label{subsec:exp_setup}

\paragraph{Model and serving backend.}
We use Qwen3-VL-32B-Instruct as the target model and implement GraphSpec by
extending \vllm v0.16.0. Experiments run on four NVIDIA A800 80GB GPUs with
tensor parallelism 4 and bfloat16 execution, with the serving engine configured
for the 4K--128K context sweep. Unless otherwise specified, the full-context
baseline, GraphSpec, and prompt-materialization baselines use the same model
weights, tokenizer, decoding configuration, software stack, and hardware environment.
GraphSpec uses the same base model for proposal and verification with $k=6$
speculative tokens. The full-context baseline and GraphSpec are evaluated on
the same complete input prompt; prompt-materialization baselines use the
shortened prompt produced by their corresponding selection strategy.

\paragraph{Dataset.}
We use SPIQA, a multimodal QA benchmark grounded in figures, tables, and text
from scientific papers. The Graph-RAG frontend processes this multimodal
evidence and returns grounded evidence passages and graph-derived priors as
inputs to the measured LLM serving stage. For long-context evaluation, we
preserve the supporting evidence and expand each context with query-specific
hard negatives.

\paragraph{Context-length control.}
To isolate the effect of context length, we do not use different datasets to
represent different context sizes. Instead, for each query, we keep the answer
and core supporting evidence fixed, and append query-specific hard-negative
passages to construct prompts of increasing length:
\[
4K,\ 8K,\ 16K,\ 32K,\ 64K,\ 128K.
\]
The distractor list for each query is fixed across all lengths; longer contexts
take progressively longer prefixes, yielding a nested length sweep in which
low-priority context increases while the core evidence remains unchanged.

\paragraph{Methods.}
We compare the following methods:
\begin{itemize}[leftmargin=*]
    \item \textbf{Full-context baseline:} standard \vllm decoding over the
    complete prompt, with the full historical \kv state available at each
    decoding step.
    \item \textbf{GraphSpec:} our runtime access planning method, which compiles
    Graph-RAG priors into runtime access plans with logical-position-preserving
    selected-\kv views and executes them through sparse self-speculative
    decoding with full-context verification.
    \item \textbf{Physical concat:} a prompt-level compression counterfactual
    that uses GraphSpec's selected evidence spans as a prompt compressor,
    materializing a short prompt instead of executing a runtime access plan.
    \item \textbf{Random concat:} a prompt-level compression baseline that
    concatenates randomly selected passages under a matched prompt budget.
    \item \textbf{Local window:} a prompt-level compression baseline that keeps
    fixed windows from the serialized context under a matched prompt budget,
    without using Graph-RAG priors.
\end{itemize}

\paragraph{Metrics.}
We report both serving efficiency and answer-level quality. Efficiency is
reported primarily through decode throughput (TPS) and time-to-first-token
(TTFT), with the main quantitative results shown in tables. For GraphSpec, we
additionally report the retained-\kv ratio $s$, the Top-3 tolerant acceptance
rate $h$, and $h-s$. For answer quality, we report LLM similarity and the
corresponding pass rate defined below. For phase-boundary analysis, we compare
the observed retained ratio, acceptance rate, and throughput speedup against
the directional implication of the cost model in
Section~\ref{sec:cost_analysis}.
All serving-path measurements are collected with diagnostic profiling disabled.
For the controlled evaluation, Graph-RAG evidence and priors are frozen before
benchmarking, so frontend retrieval is excluded from TTFT for every method.
TTFT is measured from frozen-request submission to the first generated token;
for \graphspec, it includes runtime-access-plan compilation and selected-\kv
materialization. Reported throughput covers the complete executor path rather
than isolated model-forward performance.

\paragraph{LLM-as-judge quality metric.}
Our primary open-ended QA quality metric is an LLM similarity score in $[0,1]$.
Exact string match is too brittle for our setting because correct answers may
use different wording, omit irrelevant details, or combine evidence from
multiple retrieved passages. The judge compares each generated answer with the
reference answer and produces a structured score using four dimensions: fact
consistency (40\%), information completeness (30\%), logical structure (20\%),
and expression quality (10\%). We report \emph{pass rate}, the fraction of
examples whose weighted similarity score is at least 0.5. This decision
threshold was fixed before the human audit and was not tuned using human
annotations. To independently validate the criterion, we randomly sampled 100
SPIQA answers from the evaluated outputs. Three annotators assessed answer
correctness independently without access to the corresponding LLM scores, and
their majority vote defined the human reference label. After thresholding the
LLM similarity score at 0.5, its binary decisions agreed with the majority-vote
human labels on 95 of the 100 samples. Table~\ref{tab:judge_audit} summarizes
this blinded audit. Separately, we report evidence-recall diagnostics that
measure whether the retrieved context covers gold answer, rationale, or caption
evidence, providing an orthogonal view of source coverage.

\begin{table}[t]
\caption{
\textbf{Independent human audit of the LLM-as-judge pass criterion and
evidence-coverage diagnostics.}
The 0.5 threshold was fixed before the audit. Three annotators independently
judged 100 randomly sampled SPIQA answers without access to LLM scores;
judge--human agreement compares the thresholded LLM decisions with the
majority-vote human labels. Evidence recalls measure token-level/ROUGE-1
coverage of gold answer, rationale, or caption evidence in retrieved passages.
}
\label{tab:judge_audit}
\centering
\small
\setlength{\tabcolsep}{3.2pt}
\resizebox{\linewidth}{!}{
\begin{tabular}{lccccc}
\toprule
Dataset & Samples & Judge--human agreement & Answer recall & Rationale recall & Caption recall \\
\midrule
SPIQA & 100 & 95\% & 68.8\% & 66.8\% & 76.8\% \\
\bottomrule
\end{tabular}
}
\end{table}

\subsection{RQ1: Does \graphspec decouple physical \kv consumption from serialized context growth?}
\label{subsec:rq1_scaling}

RQ1 targets the physical-consumption side of the \ksrcgap. Under standard
full-context execution, a longer serialized prompt enlarges the \kv state
consumed at every decoding step. \kap predicts a different scaling behavior:
Graph-RAG priors should let proposal-time consumption become increasingly
selective as context grows, and the avoided access should translate into
throughput gains. To test both consequences, we construct nested
evidence-preserving prompts from 4K to 128K tokens and run the full-context
baseline and \graphspec on the same complete prompt at every length.

\begin{table}[t]
\caption{
RQ1 proposal-visible \kv ratio and decode throughput over six context buckets.
All entries are averages over the evaluation set.
Active KV/source and $h$ are reported for GraphSpec; $h$ is the Top-3 tolerant
proposal acceptance rate.
}
\label{tab:rq1_scaling}
\centering
\resizebox{\linewidth}{!}{
\begin{tabular}{lrrrrrrr}
\toprule
Length & Full-context TPS & GraphSpec TPS & Speedup & Full-context TTFT & GraphSpec TTFT & Active KV/source & $h$ \\
\midrule
4K & 51.92 & 56.11 & 1.08$\times$ & 1.51s & 1.56s & 40.1\% & 0.958 \\
8K & 51.58 & 55.57 & 1.08$\times$ & 0.65s & 0.87s & 34.4\% & 0.957 \\
16K & 50.59 & 54.82 & 1.08$\times$ & 2.06s & 2.17s & 24.2\% & 0.957 \\
32K & 48.72 & 53.19 & 1.09$\times$ & 4.87s & 4.92s & 16.1\% & 0.956 \\
64K & 45.68 & 51.70 & 1.13$\times$ & 12.56s & 12.71s & 10.8\% & 0.962 \\
128K & 41.10 & 48.80 & 1.19$\times$ & 38.31s & 37.90s & 5.5\% & 0.953 \\
\bottomrule
\end{tabular}
}
\end{table}

\paragraph{Result discussion.}
Table~\ref{tab:rq1_scaling} confirms both effects. As context length increases
from 4K to 128K, the proposal-visible \kv ratio falls from 40.1\% to 5.5\%,
while proposal acceptance remains near 0.95--0.96; over the same sweep,
decode-throughput speedup rises from 1.08$\times$ to 1.19$\times$. TTFT remains
close to the full-context baseline because \graphspec preserves the complete
logical prompt and targets decode-time \kv consumption rather than prefill
elimination. The result directly supports \kap's answer to the \ksrcgap:
frontend knowledge relevance can decouple proposal-time physical consumption
from serialized context growth, and the resulting access reduction translates
into end-to-end decode gains.

\subsection{RQ2: Does \graphspec sustain efficiency gains while preserving answer quality across context lengths?}
\label{subsec:rq2_quality}

RQ2 tests whether the benefits of \kap, as instantiated by \graphspec, persist
across operating scales rather than appearing only at a favorable context
length. We combine the access and throughput results from RQ1 with answer-level
evaluation over the same 4K--128K sweep. \graphspec changes proposal-time
physical \kv consumption, while the complete logical prompt remains available
to full-context verification; a successful instantiation should therefore
improve efficiency without degrading final answers at any evaluated length.
The Top-3 tolerant acceptance rate is reported only as an execution diagnostic,
not as the quality criterion.

\begin{table}[t]
\caption{
RQ2 final-answer quality across context lengths. GraphSpec and the full-context
baseline are evaluated with the same answer-level metric; $h$ is reported only
as an execution diagnostic.
}
\label{tab:rq2_quality}
\centering
\resizebox{\linewidth}{!}{
\begin{tabular}{lrrrrr}
\toprule
Length & Full-context pass & GraphSpec pass & Full-context avg. sim. & GraphSpec avg. sim. & GraphSpec $h$ \\
\midrule
4K & 80.0\% & 80.2\% & 0.691 & 0.696 & 0.958 \\
8K & 82.6\% & 82.9\% & 0.715 & 0.716 & 0.957 \\
16K & 87.4\% & 87.4\% & 0.743 & 0.745 & 0.957 \\
32K & 86.9\% & 87.0\% & 0.744 & 0.749 & 0.956 \\
64K & 86.0\% & 87.9\% & 0.741 & 0.748 & 0.962 \\
128K & 84.6\% & 84.3\% & 0.728 & 0.732 & 0.953 \\
\bottomrule
\end{tabular}
}
\end{table}

\paragraph{Result discussion.}
Table~\ref{tab:rq2_quality} shows that \graphspec preserves answer quality
across the full context-length range. Its pass rate differs from the
full-context baseline by at most 0.3 percentage points at five of the six
lengths; at 64K it is 1.9 points higher, while average similarity remains
closely matched throughout. Thus, the efficiency gains in RQ1 are not obtained
by trading away answer quality. Combined with RQ1, where throughput improves by
1.08$\times$--1.19$\times$ and proposal-visible \kv consumption falls as
context grows, these results establish a consistent efficiency--quality
advantage across all evaluated lengths. The combined evidence validates the
\kap approach in
\graphspec: physical knowledge consumption can be reduced to deliver systems
gains without sacrificing the answer quality anchored by the complete logical
context.

\subsection{RQ3: Does the runtime access plan bridge knowledge selection and runtime consumption?}
\label{subsec:rq3_runtime_vs_prompt}

RQ3 isolates the mechanism central to \kap: whether the same frontend knowledge
is materialized as a shortened prompt or compiled into a runtime access plan.
Physical concat materializes the same evidence selected by \graphspec as a
shortened prompt, holding evidence quality fixed while removing the runtime
access plan. Random concat removes Graph-RAG-guided selection, and Local window
replaces it with a position-based heuristic; both operate under comparable
prompt budgets. If evidence selection or prompt shortening alone were
sufficient to close the \ksrcgap, these alternatives would recover
\graphspec's quality--efficiency trade-off.

\begin{table}[t]
\caption{
RQ3 ablation of evidence selection and execution strategy on 64K contexts.
``Full logical prompt'' indicates whether the original serialized context and
logical positions are preserved. For prompt-materialization baselines, accessed
KV/source is the served-prompt ratio; for GraphSpec, it is the
proposal-visible \kv ratio. Pass rate applies the pre-specified 0.5 threshold
to LLM similarity, as independently audited in Table~\ref{tab:judge_audit}.
}
\label{tab:rq3_prompt_vs_runtime}
\centering
\resizebox{\linewidth}{!}{
\begin{tabular}{llcrrrrrr}
\toprule
Method & Backend & Full logical prompt & Served prompt & Accessed KV/source & TPS & TTFT & Avg. sim. & Pass rate \\
\midrule
Full-context & standard \vllm & Yes & 65.2K & 100.0\% & 45.68 & 12.56s & 0.741 & 86.0\% \\
GraphSpec & GraphSpec & Yes & 64.9K & 10.8\% & 51.70 & 12.71s & 0.748 & 87.9\% \\
Physical concat & standard \vllm & No & 0.95K & 1.46\% & 52.82 & 1.95s & 0.653 & 73.9\% \\
Random concat & standard \vllm & No & 7.79K & 12.0\% & 52.23 & 5.17s & 0.491 & 49.7\% \\
Local window & standard \vllm & No & 7.78K & 12.0\% & 52.26 & 5.20s & 0.469 & 47.1\% \\
\bottomrule
\end{tabular}
}
\end{table}

\paragraph{Result discussion.}
Table~\ref{tab:rq3_prompt_vs_runtime} rejects that alternative. Physical concat
is the decisive ablation: despite materializing the same selected evidence in a
0.95K-token prompt, its pass rate falls to 73.9\%, compared with 87.9\% for
\graphspec. Random concat and Local window retain larger 7.8K-token prompts yet
fall further to 49.7\% and 47.1\%, confirming that arbitrary or positional
selection is not an adequate substitute for graph-derived priors. Only \graphspec
improves throughput over the full-context baseline while retaining the complete
logical context and comparable answer quality. In this controlled comparison,
the runtime access plan is therefore the operative bridge in \kap: it turns
frontend selection into backend-executable consumption, whereas evidence
selection that terminates in prompt materialization leaves the \ksrcgap
unresolved.

\subsection{RQ4: Is the observed speedup trend consistent with the phase-boundary analysis?}
\label{subsec:rq4_phase_boundary}

RQ4 evaluates the empirical consistency of the observed performance trend with
the phase-boundary analysis in Section~\ref{sec:cost_analysis}. Without
refitting the cost model to the measurements, we test its directional
implication that the positive-speedup regime becomes
more favorable as proposal acceptance remains high relative to the retained-\kv
ratio. We therefore track $s$, $h$, the margin $h-s$, and measured speedup
across context lengths.

\begin{table}[t]
\caption{
RQ4 empirical consistency with the phase-boundary model.
The retained ratio $s$ decreases as context length grows, while $h$ remains
high; consequently $h-s$ increases together with measured throughput speedup.
}
\label{tab:rq4_phase}
\centering
\begin{tabular}{lrrrr}
\toprule
Length & $s$ & $h$ & $h-s$ & Actual speedup \\
\midrule
4K & 0.401 & 0.958 & 0.557 & 1.08$\times$ \\
8K & 0.344 & 0.957 & 0.614 & 1.08$\times$ \\
16K & 0.242 & 0.957 & 0.715 & 1.08$\times$ \\
32K & 0.161 & 0.956 & 0.796 & 1.09$\times$ \\
64K & 0.108 & 0.962 & 0.853 & 1.13$\times$ \\
128K & 0.055 & 0.953 & 0.897 & 1.19$\times$ \\
\bottomrule
\end{tabular}
\end{table}

\paragraph{Result discussion.}
Table~\ref{tab:rq4_phase} is consistent with the phase-boundary prediction. As
contexts grow, $s$ decreases from 0.401 to 0.055 while $h$ remains high, so the
margin $h-s$ increases from 0.557 to 0.897. Measured speedup follows the same
direction, rising from 1.08$\times$ to 1.19$\times$. The agreement supports the
directional implication of the phase-boundary model: a larger \kv-side
acceptance--retention margin corresponds to a more favorable regime for
plan-guided execution. It thereby closes the \kap
analysis--system--experiment loop by connecting the compiler's planning
criterion, the executor's observed behavior, and end-to-end speedup.

% =============================================================================

\section{Related Work}
\label{sec:related_work}

\paragraph{Retrieval-augmented generation and graph-guided retrieval.}
Retrieval-augmented generation grounds LLM outputs in external knowledge by retrieving relevant passages or documents before generation \citep{lewis2020retrieval,guu2020realm}. Generative retrieve-and-read systems such as Fusion-in-Decoder and retrieval-enhanced language models such as RETRO integrate retrieved passages into generation \citep{izacard2021leveraging,borgeaud2022improving}. Dense retrieval methods such as DPR improve open-domain retrieval by learning neural passage representations \citep{karpukhin2020dense}, and multi-hop retrieval extends dense selection to complex questions requiring evidence chains \citep{xiong2021multi}. Graph-based retrieval and reasoning methods build structured representations over entities, documents, or events for multi-hop QA \citep{tu2019multi,edge2024from,gutierrez2024hipporag}. These works improve the \emph{frontend knowledge selection} stage. KAP is complementary: it asks how the selected knowledge should be consumed by the LLM serving backend once it has been retrieved.

\paragraph{RAG serving and KV-cache reuse.}
General LLM serving systems improve batching, scheduling, offloading, and cache
reuse for autoregressive generation
\citep{yu2022orca,aminabadi2022deepspeed,sheng2023flexgen,kwon2023efficient,zheng2024sglang}.
Recent RAG-specific systems optimize serving by caching or reusing intermediate
states of retrieved knowledge. RAGCache caches retrieved knowledge states in a
hierarchy to reduce TTFT and improve throughput \citep{jin2024ragcache}.
CacheBlend fuses cached knowledge chunks even when they are not strict prefixes
of the input \citep{yao2024cacheblend}, and TurboRAG precomputes KV caches for
document chunks to accelerate RAG prefill \citep{lu2025turborag}. RAGO studies
systematic performance optimization for different RAG pipelines through a
structured RAGSchema abstraction \citep{jiang2025rago}. These systems reduce
retrieval-generation latency, especially around prefill, scheduling, and cache
reuse. In contrast, KAP addresses the runtime-consumption side of
knowledge-augmented inference: frontend knowledge priors are compiled into
executable access plans so that the backend can selectively consume knowledge
rather than uniformly processing the serialized prompt. Unlike pipeline
schemas that organize retrieval and generation components, the runtime access
plan is a backend-consumed IR that governs physical knowledge access during
execution.

\paragraph{Prompt compression and long-context usage.}
Prompt compression methods reduce inference cost at the input layer by
shortening the serialized context before inference. LLMLingua and
LongLLMLingua compress prompts to reduce cost and latency while attempting to
preserve task-relevant information
\citep{jiang2023llmlingua,jiang2024longllmlingua}; Selective Context similarly
prunes redundant context for efficiency \citep{li2023selective}. These methods
are complementary to KAP: they transform the prompt presented to the model,
whereas KAP preserves the logical input and moves selection into an executable
runtime access plan. The distinction matters in long-context and graph-based
QA settings, where models can be sensitive to the position of relevant
information \citep{liu2024lost} and where evidence boundaries, provenance, or
graph organization may carry reasoning value \citep{tu2019multi,xiong2021multi}.
GraphSpec therefore treats frontend priors not merely as prompt-construction
hints, but as runtime guidance for physical \kv consumption.

\paragraph{KV-cache compression and sparse attention.}
A large body of work reduces the cost of long-context inference by optimizing
attention or compressing KV access. IO-aware kernels such as FlashAttention
reduce attention memory traffic \citep{dao2022flashattention}. H$_2$O keeps
heavy-hitter tokens in the KV cache \citep{zhang2023h2o}; StreamingLLM
identifies attention sinks for stable streaming inference
\citep{xiao2024streamingllm}; SnapKV and PyramidKV compress KV states by
selecting important tokens or allocating cache across layers
\citep{li2024snapkv,cai2024pyramidkv}. Quest performs query-aware KV-page
selection at inference time \citep{tang2024quest}, while MInference accelerates
long-context prefill through dynamic sparse attention patterns
\citep{jiang2024minference}. These methods infer token or page importance from
model-internal attention or query statistics. KAP instead uses
\emph{frontend knowledge priors} as the source of runtime access decisions, and
GraphSpec applies sparse access only to the proposal pass while using
full-context verification to guard answer-level quality. Accordingly, our
evaluation isolates this orthogonal systems mechanism rather than providing an
exhaustive ranking of general-purpose \kv-cache compression methods.

\paragraph{Speculative and self-speculative decoding.}
Speculative decoding accelerates generation by drafting candidate tokens with
a cheaper path and verifying them with a target model
\citep{leviathan2023fast,chen2023speculative}. System-oriented extensions such
as SpecInfer organize candidates into token trees for speculative verification
\citep{miao2024specinfer}. Subsequent approaches improve drafting through
multiple decoding heads, feature-level prediction, early exits, or parallel
decoding
\citep{cai2024medusa,li2024eagle,elhoushi2024layerskip,fu2024lookahead}.
Long-context variants pursue several complementary routes: TriForce performs
hierarchical speculation with dynamic sparse-\kv retrieval
\citep{sun2024triforce}; RAPID drafts from shortened retrieval contexts
\citep{chen2025rapid}; QuantSpec uses hierarchically quantized weights and
\kv states for self-speculation \citep{tiwari2025quantspec}; and LongSpec
employs a learned memory-efficient drafter with a constant-sized \kv cache
\citep{yang2026longspec}. GraphSpec adopts selected-\kv proposal with
full-context verification as one concrete \kap executor policy. Its distinction
lies in how the proposal view is produced: Graph-RAG priors are compiled into
a runtime access plan and a logical-position-preserving selected-\kv view,
rather than being materialized as a shortened prompt or inferred solely within
the drafting mechanism.

% =============================================================================

\section{Limitations}
\label{sec:discussion}

Our evaluation validates \kap through \graphspec, one concrete instantiation built
with a Graph-RAG knowledge-selection frontend and a \vllm serving backend.
Therefore, our empirical evidence does not exhaust the broader design space of
runtime knowledge consumption. The reported results are strongest for
long-context QA workloads where graph-derived priors can be grounded in evidence
spans and realized as backend runtime objects. Other \kap instantiations may
use different frontends, plan compilers, access policies, verification
mechanisms, or LLM serving backends, and their effectiveness will depend on the
quality of the knowledge priors, the compiler's ability to translate them into
useful runtime access plans, and the backend's ability to execute those plans
without excessive metadata, scheduling, or synchronization overhead. Future
\kap instantiations could realize runtime access plans through alternative
sparse-access, tiered-\kv, or disaggregated execution mechanisms. Thus, our
results establish \kap as a feasible serving paradigm and identify a broader
design space for future runtime knowledge-consumption systems.

% =============================================================================

\section{Conclusion}
\label{sec:conclusion}

We identified the \ksrcgap as an architectural mismatch in
knowledge-augmented LLM serving: knowledge-selection frontends produce high-value priors
about relevant knowledge, yet conventional LLM serving backends consume only the
serialized prompt and its full \kv state. To bridge the gap, we introduced
\kap, a paradigm for plan-driven knowledge execution organized around a
universal serving-time IR that compiles structured knowledge priors into physical-access
policies while preserving logical prompt semantics. \graphspec provides a
reference implementation with a Graph-RAG knowledge-selection frontend and a
\vllm serving backend. Our phase-boundary analysis and controlled long-context
QA evaluation connect knowledge selectivity, runtime consumption, and systems
efficiency. Beyond \graphspec, \kap provides a universal execution substrate
for knowledge-selection--execution co-design across multimodal, agentic, and deliberative
long-context systems, where semantic knowledge selection and physical context
consumption must be optimized jointly.

% =============================================================================
% References.
% =============================================================================
\bibliography{references}
\bibliographystyle{iclr2025_conference}

\end{document}